\documentclass[runningheads]{llncs}
\usepackage[T1]{fontenc}
\usepackage{graphicx}
\usepackage{cite}
\usepackage{amsmath} 
\usepackage{amssymb}
\usepackage{url} 
\usepackage{multirow}
\usepackage{booktabs}
\usepackage{multirow}
\usepackage{lscape}
\usepackage{subcaption}
\usepackage{comment}
\usepackage[utf8]{inputenc}
\usepackage[T1]{fontenc}
\usepackage{lineno}
\usepackage{float}
\usepackage{xcolor}
\usepackage{soul}
\definecolor{light-gray}{gray}{0.8} 
\usepackage{multirow}
\usepackage{colortbl}

\usepackage{color}
\usepackage{silence}
\begin{document}
\title{Adaptive Machine Learning for Resource-Constrained Environments}
%
%
\author{Sebastián Andrés Cajas\inst{1}\orcidID{0000-0003-0579-6178} \and
Jaydeep Samanta\inst{1}\orcidID{0009-0006-5716-5545} \and
Andrés L. Suárez-Cetrulo\inst{1}\orcidID{0000-0001-5266-5053}\and 
Ricardo Simón Carbajo \orcidID{0000-0002-2121-2841}
}
\authorrunning{Cajas, Samanta, Suárez-Cetrulo and Simon Carbajo}
%
\institute{Ireland’s Centre for Artificial Intelligence (CeADAR), University College Dublin, D04 V2N9, Dublin, Ireland. }
\maketitle              

\def\thefootnote{}\footnotetext{
All authors have equally contributed to the development of this work.}

\begin{abstract}
The Internet of Things is an example domain where data is perpetually generated in ever-increasing quantities, reflecting the proliferation of connected devices and the formation of continuous data streams over time. Consequently, the demand for ad-hoc, cost-effective machine learning solutions must adapt to this evolving data influx. 
This study tackles the task of offloading in small gateways, exacerbated by their dynamic availability over time. An approach leveraging CPU utilization metrics using online and continual machine learning techniques is proposed to predict gateway availability. These methods are compared to popular machine learning algorithms and a recent time-series foundation model, Lag-Llama, for fine-tuned and zero-shot setups. 
Their performance is benchmarked on a dataset of CPU utilization measurements over time from an IoT gateway and focuses on model metrics such as prediction errors, training and inference times, and memory consumption. Our primary objective is to study new efficient ways to predict CPU performance in IoT environments. 
Across various scenarios, our findings highlight that ensemble and online methods offer promising results for this task in terms of accuracy while maintaining a low resource footprint. 


\end{abstract}

\section{Introduction}
\label{sec:intro}

In today's dynamic world, data streams are abundant and ever-changing, spanning non-stationary environments where data evolves over time. Machine learning (ML) models in these domains may need regular model updates to minimize the degradation of their performance over time, as seen in weather prediction and customer preference model \cite{tsymbal2004problem},  social networks, sensor networks, and financial data streams \cite{SUAREZCETRULO2023118934}. The growing volume of data generated underscores a pressing need for real-time processing capabilities. This perpetual evolution in data undermines model predictions, as outdated data distributions may no longer align with current data, necessitating frequent model updates and introducing the challenge of concept drift in data streams \cite{tsymbal2004problem}. Such limitations can impede AI systems, rendering them incapable of effectively adapting to ongoing changes and struggling with memory constraints to process incoming data \cite{gomes2017adaptive}. Furthermore, the evolving dynamics of the data streams, whereby behaviors may also re-occur, make it necessary to consider forgetting mechanisms \cite{gama2013evaluating} and reusing former active learners \cite{SUAREZCETRULO2023118934}. 

CPU demand is a primary driver of resource shortages in virtualized environments, significantly impacting host-machine performance \cite{zhang2008predict, bey2009cpu}. 
Accurately predicting future resource usage for impending demands stands as one of the significant challenges in cloud computing \cite{mason2018predicting}, which is particularly challenging due to the non-stationarity of CPU utilization and the potential presence of concept drifts \cite{duggan2017predicting}. This can result in inefficient resource allocation across machines. Thus, forecasting CPU allocation accurately can help reduce energy consumption~\cite{janardhanan2017cpu}. 
Such non-stationarity may arise from many background processes tracing periodic and non-periodic behavior with sudden peaks of loads \cite{karim2021bhyprec}. Hence, estimating CPU utilization levels can be crucial in aligning tasks with resources, maximizing their availability, and minimizing computational costs \cite{valarmathi2021resource}. 
 
Traditionally, statistical predictive models such as Autoregressive Integrated Moving Average (ARIMA) and family variations have been focused on optimizing the cost functions \cite{makridakis2018statistical}, allowing a good fit to the data but limiting their adaptability for non-linear-trends and long-term dependencies \cite{borovykh2018dilated, patel2022hybrid}. More recently, neural networks have shown stronger capabilities to fill that gap; for example, one-dimensional Convolutional Neural Networks (CNN) have shown their effectiveness for pattern extraction on 1-dimensional complex signals \cite{kiranyaz20211d} and nonlinearity time-series extraction \cite{wang2017time}. Similarly, as introduced Long Short-Term Memory (LSTM) networks \cite{hochreiter1997long} to cover the vanishing gradient problem of Recurrent Neural Networks (RNN) which has allowed neural networks to learn longer-term dependencies that have extended over all LSTM-based models \cite{valarmathi2021resource, karim2021bhyprec} outperforming traditional methods \cite{hsu2016bridging}. 
Finally, online incremental ML algorithms allow drift handling in data streams with an efficiency that suits resource-aware environments \cite{SUAREZCETRULO2023118934}. This allows for quick adjustments to temporal changes, largely owing to the incorporation of forgetting mechanisms, ensuring rapid adaptation to new patterns~\cite{gama2013evaluating}. 

This paper aims to be a comparative study of the performance of the online regression models 
on a novel application dataset of CPU loads that exhibits non-stationary patterns over time. 
Our study delves into a comparative analysis of various classic ML models alongside online learning algorithms, assessing accuracy and computation performance metrics. 
Furthermore, these models are juxtaposed with recent deep learning methods and the time-series foundation model Lag-Llamma~\cite{rasullag}. 
This research aims to contribute to the advancement of IoT systems by providing insights into model selection for CPU performance estimation through the use of online ML and other modern algorithms. This work also provides insights into the suitability of online regression models in our application domain. 
The main contributions of this paper are outlined below.

\begin{enumerate}
    \item \textit{CPU utilization prediction}: This paper proposes an approach to predicting CPU load in IoT gateways using state-of-the-art ML algorithms. 
    While offline ensemble methods offer the best trade-off between accuracy and computational cost, continual learning methods offer promising results in predicting CPU loads accurately for edge devices. 
    \item \textit{Evaluation benchmark:} A benchmark is proposed to compare traditional versus online and foundation models.  
    In addition to performance evaluation, the memory and runtime of the models are computed as a measure of their footprint. 
    This assessment allows the identification of the most effective model for CPU performance estimation and considers the chosen approach's computational and environmental implications. 
    \item \textit{Code and data sharing}: The code and data generation used in our experiments are publicly available~\footnote{GitHub repository: \url{https://github.com/sebasmos/AML4CPU}} to facilitate reproducible research and encourage collaboration in the research community. Researchers and practitioners can leverage this codebase to replicate our findings and build upon our work.
\end{enumerate}

The paper is structured as follows: After the introduction, the first section covers related work for CPU utilization prediction and the techniques used in this paper. Next, the research data section provides a description of the dataset proposed. Subsequently, this paper presents the experimental section, outlining the methodology followed, metrics used, models, and discussion of results obtained.   
Finally, conclusions and future lines of work are drawn. 
\section{Related work} 

Predicting CPU utilization has been approached through different methods, including more traditional methods such as polynomial fitting \cite{zhang2008predict}, regression-based models such as linear regression \cite{farahnakian2016energy}, and gradient-descent optimizers like stochastic gradient descend (SGD)\cite{ruder2016overview}. 
Other advanced methods include adaptive networks with clustering \cite{bey2009cpu} and stack generalization, which combines algorithms such as KNN and decision trees (DT). 

Shaikh et al. \cite{shaikh2024prediction} used DTs \cite{breiman2017classification} to forecast CPU usage in VM workloads. DTs start learning by splitting the first node based on a metric such as information gain or the Gini coefficient \cite{cetrulo2022adaptive}. This split triggers the creation of new nodes, which may split again during the learning process. Final nodes without children predict outcomes for both classification and regression tasks. 

Based on base models such as DTs, ensemble methods can be constructed by aggregating multiple predictive models (weak learners) for improved accuracy, relying on a voting mechanism. These have recently been used for VM resource allocation \cite{rahmanian2018learning}. 
Some example methods are Adaboost, XGBoost \cite{chen2016xgboost}, and random forests (RF). 
RF \cite{iqbal2019adaptive} specifically builds upon DTs by training multiple decision trees on different data subsets to promote diversity and make predictions based on the majority vote, enhancing accuracy and stability. 

Recent work on predicting data center workloads includes Kim et al.'s study \cite{kim2018cloudinsight}, which combines Linear Regression, support vector machines, and time-series models with dynamic weight adjustment. 
Additionally, support vector regression \cite{drucker1996support} and Kalman smoothing \cite{hu2013cpu} have demonstrated effectiveness in handling dynamic characteristics for accurate predictions on CPU load and cloud prediction.
Another incremental approach worth exploring for this task is the Passive-Aggressive algorithm (PA) \cite{crammer2006online}, which adapts the model based on feedback and can help with the dynamic nature of the CPU load changes. This was initially proposed for binary classification, incrementally updating the decision boundary, and later extended to regression tasks.\color{black}

Neural-network-based approaches \cite{mason2018predicting, duggan2017predicting, janardhanan2017cpu, karim2021bhyprec}, LSTMs~\cite{hochreiter1997long},  and hybrid models combining ensembling models with LSTMs \cite{valarmathi2021resource}
have also recently been used for CPU utilization prediction. LSTMs are a type of RNN designed to address the vanishing gradient problem that affects standard RNNs and as potent tools for processing and forecasting time series data across diverse domains  \cite{moghar2020stock,mason2018predicting}. 
Mason et al. (2018) specifically explored the potential of neural networks in CPU utilization forecasting, developing evolutionary neural networks through an evolutionary optimization algorithm. 
Moreover, LSTMs have been employed in CPU utilization forecasting, often compared against traditional techniques like ARIMA \cite{janardhanan2017cpu}. Additionally, various architectures such as Recurrent Neural Networks (RNNs), Bidirectional LSTMs (BiLSTMs), and hybrid versions like BiLSTM-RNN models and CNN-LSTM \cite{patel2022hybrid} have demonstrated applicability in this domain \cite{karim2021bhyprec, patel2022hybrid}. 

Hoeffding Trees (HT) \cite{domingos2000mining} were originally designed for constructing and updating decision trees in dynamic data streams. Leveraging the Hoeffding bound, a statistical inequity, they efficiently determine optimal splits at each node without requiring full dataset analysis, thus becoming highly memory-efficient. The Hoeffding Adaptive Tree (HAT) enhances adaptiveness by replacing old branches dynamically using metrics such as Adaptive Windowing (ADWIN) algorithm \cite{bifet2009adaptive} and also proposes a bootstrapping sampling on top of Hoeffding Trees. 
Bagging \cite{bifet2010leveraging} and boosting-based \cite{chen2012online} techniques 
have recently proven their success as part of ensembles in data stream learning like Adaptive Random Forests (ARF) \cite{gomes2017adaptive} and Streaming Random Patches (SRP) \cite{gomes2019streaming}. 
ARF \cite{gomes2017adaptive} is an enhanced adaptive ensemble with diversity through resampling and random node splitting, equipped with drift detection per node for adaptive training. ARF uses enhanced HTs as base learners and ADWIN as a drift detector to understand when to train and replace decision trees. 

Finally, the advent of foundation models in artificial intelligence has created a trend for reusing pre-trained models, something already common in the data stream learning field~\cite{SUAREZCETRULO2023118934}. 
Lag-Llama has recently emerged as a time-series foundational model  \cite{rasullag}, leveraging the properties of the decoder-only
transformer-based architecture LLaMA and incorporating pre-normalization via the RMSNorm. A current topic of discussion in foundation models, which tend to be multi-purpose and thus experiment domain drifts over time, is the issue of alignment and models sharing a similar world representation. This topic has an analogy in data stream learning, as models representing similar data distributions are often contrasted with each other in the meta-learning field, comparing the similarity of the data fed to them (\textit{concept similarity}) or their predictive results (\textit{conceptual equivalence}) \cite{SUAREZCETRULO2023118934}. 

Predicting CPU performance efficiently is crucial for optimizing system resources and enhancing overall computational efficiency. In this comparative study, we delve into the performance evaluation of state-of-the-art classical models, deep learning, online ML, and a time-series foundational in both zero-shot and fine-tuned setups for this predictive task.
Our research endeavors to contribute to the advancement of new methodologies for CPU performance estimation tasks and offer a new dataset for data stream learning.

\section{Research Data}

The hardware utilized for data collection was an Orange Pi 5\footnote{\url{http://www.orangepi.org/html/hardWare/computerAndMicrocontrollers/details/Orange-Pi-5.html}}, powered by the 8-core RK3588S processor. 

The data collection was performed using the \textit{psutil} library, recording CPU usage per core and UNIX timestamps at 1-minute intervals. 
This process ran over $\approx$ 32 days (47,315 minutes) while subjecting the system to a \textit{stress-ng test}, which simulates diverse workloads, engaging all CPU cores at varying utilization levels (0-100\%) through random generation. This used workloads of 60 minutes followed by a 60-second pause before initiating the next test. 
To isolate CPU behavior, the \textit{stress-ng test} was configured to focus exclusively on CPU usage. 

The collected samples underwent a resampling process to ensure an evenly distributed index with precisely one-minute intervals between each sample.  
The 47,315 data samples were partitioned into 37,852 samples (80\%) for training and 9,463 (20\%) for testing.
The datasets used in this paper are publicly available in the data folder of our GitHub repository\footnote{https://github.com/sebasmos/AML4CPU/tree/main/data}. This includes training and testing sets, named $train\_data.csv$ and $test\_data.csv$, respectively. 

The feature set used in the experiments is univariate, using lags of CPU utilization to predict the next one; thus, models in this paper will provide 1-minute ahead predictions. Different experiments have used different window sizes (WS) or lags lengths \( L \), where \( L \) refers to the past CPU utilization measurements \( x(t-1) \), \( x(t-7) \), \( x(t-14) \),..., \( x(t-L) \), where  \( L \)  is the maximum lag index used in the model.  The target feature is the CPU utilization one step ahead,  \( x(t) \).

\section{Experiments}

Three distinct experiments were carried out to evaluate the optimal models using the CPU dataset over 20 different seeds. 

\begin{itemize}
    \item \textit{Experiment I}: A hold-out benchmarking process was conducted between state-of-the-art ML algorithms. 
    \item \textit{Experiment II}: Online incremental learners were evaluated using the training and test sets from Experiment I for pre-training and for a prequential evaluation~\cite{SUAREZCETRULO2023118934} respectively. 
    \item \textit{Experiment III}: A zero-shot and fine-tuning setup of the time-series foundation model Lag-Llama was run as in the previous experiments to compare the generalization capabilities of foundation models against other state-of-the-art and online ML methods. 
\end{itemize}

All experiments were performed in plain vanilla settings. The only parameters tweaked were the \textit{window size}, which relates to the length of the feature set, and the \textit{recommended values for Lag-Llama}: context length and RoPE. This was both zero-shot and fine-tuned with the training set. 
Each experiment was run 20 times with different seeds to handle non-deterministic models, providing mean and standard deviation across runs and boxplots for them. The libraries used for this study were \textit{river} for online ML, \textit{scikit-learn} for classical methods, and \textit{PyTorch} for deep learning. 
Detailed results are provided in Tables \ref{tab:exp1_full_table}, \ref{tab:exp2_full_table}, and \ref{tab:exp3_finetuned_lagllama}. 
In these tables, we highlight the best results for each algorithm across different WSs marked in bold. We will focus on these bold results for analysis, with the overall best results in each experiment marked in gray. Boxplots and scatterplots exhibiting similar patterns in this experimental section are also excluded to simplify the analysis.

To assess model performance, this work employed a variety of error metrics \cite{botchkarev2018performance}. These are covered below with their mathematical intuition. N represents the number of data points in the dataset, $y_i$ represents the actual value, and $\hat{y}i$ represents the predicted value of i$^{th}$ data point in the dataset.

\begin{itemize}
    \item \textit{Mean Absolute Error (MAE)} is computed by the average of the absolute difference between the predicted and actual values: 
    $MAE(y, \hat{y}) = \frac{\sum_{i=0}^{N - 1} |y_i - \hat{y}_i|}{N}$. 
    \item \textit{Mean Squared Error (MSE)} measures the average squared difference between the actual and predicted values:
    $MSE(y, \hat{y}) = \frac{\sum_{i=0}^{N - 1} (y_i - \hat{y}_i)^2}{N}$
    \item \textit{Root Mean Squared Error (RMSE)} is the square root of the MSE and can be represented as $RMSE(y, \hat{y}) = \sqrt{\frac{\sum_{i=0}^{N - 1} (y_i - \hat{y}_i)^2}{N}}$. 
    \item \textit{Mean Absolute Percentage Error (MAPE)} measures the average absolute percentage difference between the actual and predicted values: 
    $MAPE(y, \hat{y}) = \frac{100\%}{N} \sum_{i=0}^{N - 1} |\frac{y_i - \hat{y}_i}{y_i}|$.
    \item Symmetric mean absolute percentage error (SMAPE) is introduced to overcome the asymmetric nature of MAPE:  
    $SMAPE(y, \hat{y}) = \frac{100\%}{N} \sum_{i=0}^{N - 1} \frac{ 2*|y_i - \hat{y}_i|}{|y| + |\hat{y}|}$.
    \item \textit{Mean Absolute Scaled Error (MASE)} is determined by calculating the mean absolute error of actual forecasts and the mean absolute error produced by a naive forecast calculated using the in-sample data. 
    $MASE = \frac{MAE}{MAE_{in-sample,naive}}$
    \item \textit{R-squared error} (\( R^2 \)) measures the percentage of the target variable's overall variance that can be accounted for by the model's predictions: 
    $R^2 (y, \hat{y}) = 1 - \frac{\sum_{i=1}^{N} (y_i - \hat{y}_i)^2}{\sum_{i=1}^{N} (y_i - \bar{y})^2}$.
\end{itemize}

In addition, measurements of training, evaluation time, and memory consumption per model using \textit{asizeof.asizeof(model)} were captured to understand the model's footprints. Training and evaluation times were measured in seconds, and memory was measured in megabytes (MB). 
These experiments have been run in a server with 32-core AMD Ryzen Threadripper PRO 5975WX, 256 GB of RAM, and 2 x NVIDIA GeForce RTX 4090 GPUs. The GPU has mainly been used for Lag-LLama in Experiment III, while the rest of the algorithms have been run in CPU to allow comparable runtimes. 

\subsection{Experiment I}

\begin{figure}[ht]
  \centering
  \includegraphics[width=\linewidth]{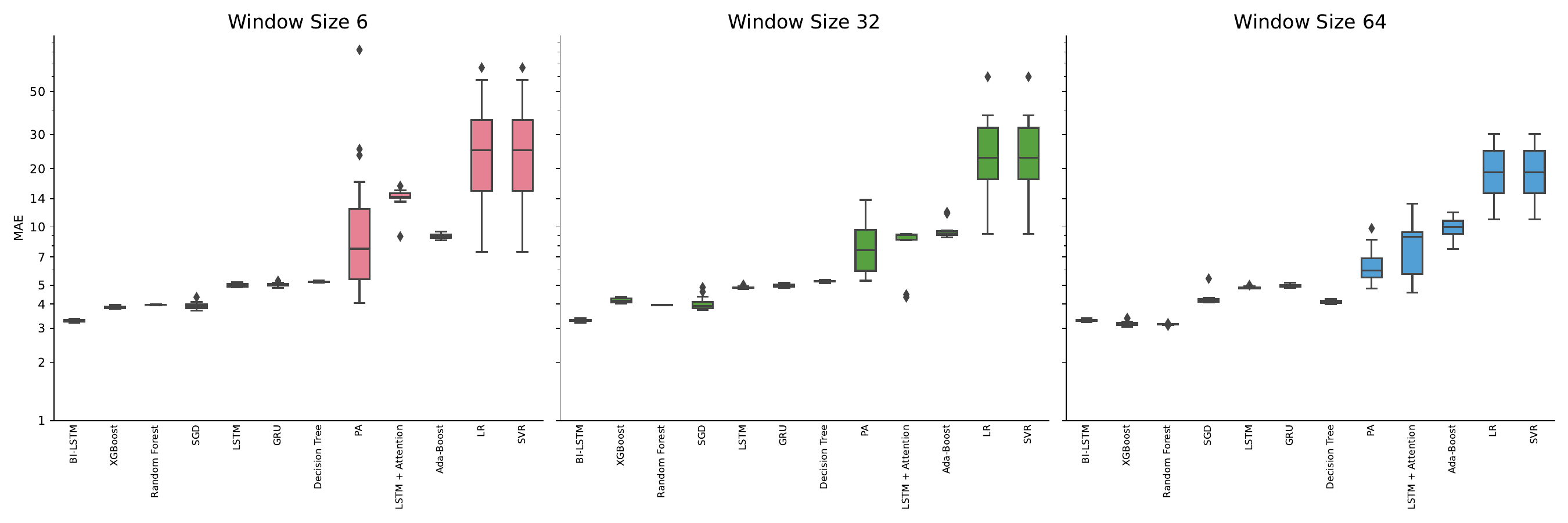}
  \caption{MAE per model in Experiment I at different window sizes.}
  \label{fig:box_plot_exp1}
\end{figure}

\begin{figure}[ht!]
  \centering
  \begin{subfigure}[b]{\linewidth}
  \includegraphics[width=\linewidth]{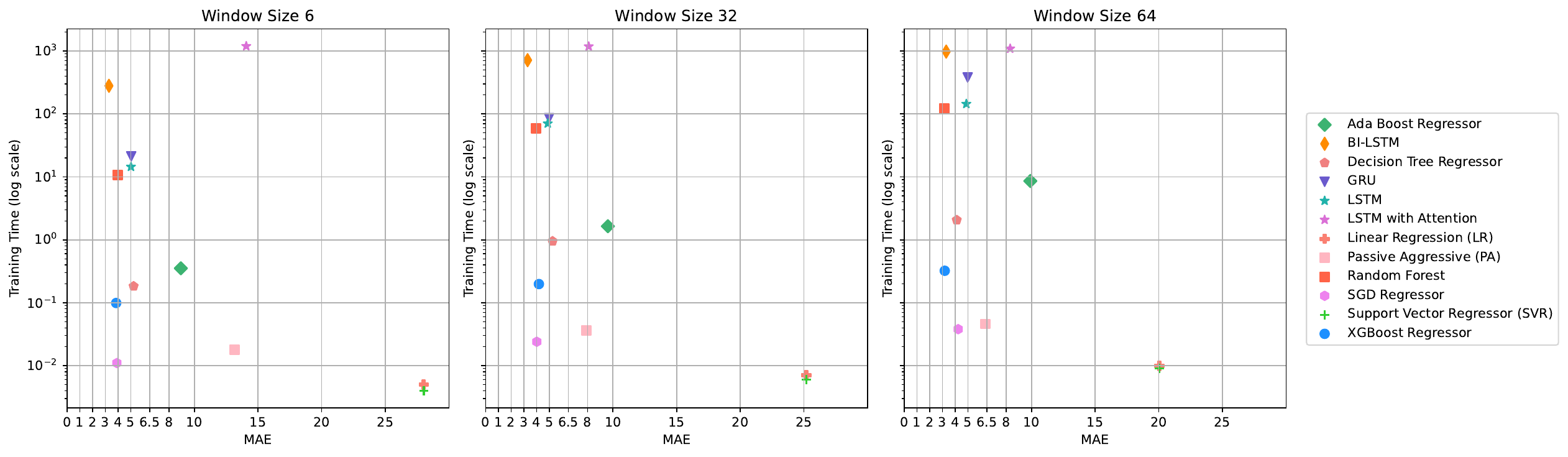}
  \end{subfigure}
      \caption{Training time vs. MAE per model in Experiment I.}
  \label{fig:exp1_scaterplots}
\end{figure}

Firstly, state-of-the-art ML models are compared as detailed in Table \ref{tab:exp1_full_table}. 
Subsequently, we evaluate their performance using hold-out validation. 
The outcomes are visually represented through model boxplots and scatterplots in Figures~\ref{fig:box_plot_exp1}~and~\ref{fig:exp1_scaterplots}.

\begin{table}[ht!]
\centering

\caption{Experiment I with highlighted results for WS with the lowest MAE across 20 runs. Values are rounded to a maximum of three decimal places.}

\resizebox{1.0\columnwidth}{!}{%
\label{tab:exp1_full_table}
\centering

\hskip-0.0cm\begin{tabular}{lllllllllllllll}
\hline
\multirow{2}{*}{\textbf{Model}} & \multirow{2}{*}{\textbf{WS}} & \multicolumn{2}{l}{\textbf{MAE}} & \multicolumn{2}{l}{\textbf{RMSE}} & \multicolumn{2}{l}{\textbf{SMAPE}} & \multicolumn{2}{l}{\(\mathbf{R^2}\)} & \multicolumn{2}{l}{\textbf{MASE}} & \textbf{Train time (s)} & \textbf{Evaluation (s)} & \textbf{Memory } \\ \cline{3-15} 
 &  & \textbf{mean} & \textbf{std} & \textbf{mean} & \textbf{std} & \textbf{mean} & \textbf{std} & \textbf{mean} & \textbf{std} & \textbf{mean} & \textbf{std} & \textbf{mean} & \textbf{mean} & \textbf{mean} \\ \hline 
\multirow{6}{*}{\textbf{XGBoost Regressor}} & 6 & 3.845 & 0.056 & 9.391 & 0.065 & 22.229 & 0.318 & 0.903 & 0.001 & 0.994 & 0.014 & 0.099 & 0.003 & 0.004 \\
 & 9 & 3.902 & 0.079 & 9.408 & 0.082 & 22.231 & 0.439 & 0.902 & 0.002 & 1.008 & 0.02 & 0.11 & 0.003 & 0.004 \\
 & 12 & 3.963 & 0.089 & 9.472 & 0.095 & 22.413 & 0.544 & 0.901 & 0.002 & 1.024 & 0.023 & 0.123 & 0.002 & 0.004 \\
 & 20 & 4.045 & 0.107 & 9.474 & 0.104 & 23.037 & 0.493 & 0.901 & 0.002 & 1.044 & 0.028 & 0.158 & 0.003 & 0.004 \\
 & 32 & 4.178 & 0.111 & 9.553 & 0.113 & 23.329 & 0.468 & 0.899 & 0.002 & 1.078 & 0.029 & 0.198 & 0.013 & 0.004 \\
 \rowcolor{light-gray}
 &  \textbf{64} & \textbf{3.185} & \textbf{0.103} & \textbf{7.344} & \textbf{0.424} & \textbf{21.881} & \textbf{0.381} & \textbf{0.941} & \textbf{0.007} & \textbf{0.822} & \textbf{0.027} & \textbf{0.322} & \textbf{0.01} & \textbf{0.004} \\ \hline
\multirow{6}{*}{\textbf{Ada Boost Regressor}} & 6 & 8.933 & 0.23 & 12.358 & 0.244 & 32.739 & 0.626 & 0.831 & 0.007 & 2.308 & 0.06 & 0.35 & 0.003 & 0.013 \\
 & \textbf{9} & \textbf{8.901} & \textbf{0.238} & \textbf{12.381} & \textbf{0.3} & \textbf{32.699} & \textbf{0.637} & \textbf{0.83} & \textbf{0.008} & \textbf{2.303} & \textbf{0.061} & \textbf{0.449} & \textbf{0.003} & \textbf{0.013} \\
 & 12 & 9.266 & 0.189 & 12.969 & 0.175 & 33.432 & 0.48 & 0.814 & 0.005 & 2.393 & 0.049 & 0.745 & 0.004 & 0.015 \\
 & 20 & 10.32 & 1.814 & 14.282 & 2.191 & 35.313 & 3.275 & 0.77 & 0.075 & 2.664 & 0.468 & 1.154 & 0.005 & 0.015 \\
 & 32 & 9.609 & 0.974 & 13.713 & 1.185 & 34.35 & 1.793 & 0.791 & 0.039 & 2.479 & 0.251 & 1.685 & 0.005 & 0.014 \\
 & 64 & 9.765 & 1.114 & 12.932 & 1.009 & 36.262 & 2.16 & 0.815 & 0.029 & 2.521 & 0.288 & 8.665 & 0.021 & 0.04 \\ \hline
\multirow{6}{*}{\textbf{Decision Tree Regressor}} & 6 & 5.221 & 0.033 & 13.468 & 0.105 & 29.861 & 0.261 & 0.8 & 0.003 & 1.349 & 0.008 & 0.184 & 0.003 & 0.002 \\
 & 9 & 5.289 & 0.047 & 13.621 & 0.14 & 29.54 & 0.168 & 0.795 & 0.004 & 1.366 & 0.012 & 0.272 & 0.003 & 0.002 \\
 & 12 & 5.271 & 0.052 & 13.702 & 0.169 & 28.972 & 0.184 & 0.793 & 0.005 & 1.361 & 0.014 & 0.362 & 0.003 & 0.002 \\
 & 20 & 5.194 & 0.05 & 13.383 & 0.136 & 28.793 & 0.359 & 0.802 & 0.004 & 1.341 & 0.013 & 0.599 & 0.003 & 0.002 \\
 & 32 & 5.254 & 0.045 & 13.407 & 0.131 & 30.717 & 0.195 & 0.802 & 0.004 & 1.355 & 0.012 & 0.966 & 0.003 & 0.002 \\
 & \textbf{64} & \textbf{4.119} & \textbf{0.074} & \textbf{9.783} & \textbf{0.294} & \textbf{26.17} & \textbf{0.177} & \textbf{0.895} & \textbf{0.006} & \textbf{1.065} & \textbf{0.019} & \textbf{2.065} & \textbf{0.003} & \textbf{0.003} \\ \hline
\multirow{6}{*}{\textbf{Random Forest Regressor}} & 6 & 3.967 & 0.012 & 9.407 & 0.014 & 22.216 & 0.11 & 0.902 & 0.001 & 1.025 & 0.003 & 10.811 & 0.168 & 0.086 \\
 & 9 & 3.932 & 0.013 & 9.308 & 0.021 & 21.56 & 0.11 & 0.904 & 0.001 & 1.016 & 0.003 & 16.121 & 0.169 & 0.083 \\
 & 12 & 3.924 & 0.012 & 9.281 & 0.021 & 21.33 & 0.127 & 0.905 & 0.001 & 1.014 & 0.003 & 21.552 & 0.168 & 0.083 \\
 & 20 & 3.907 & 0.011 & 9.269 & 0.02 & 21.152 & 0.11 & 0.905 & 0.001 & 1.009 & 0.003 & 36.062 & 0.168 & 0.083 \\
 & 32 & 3.95 & 0.012 & 9.348 & 0.032 & 21.555 & 0.129 & 0.904 & 0.001 & 1.019 & 0.003 & 58.982 & 0.169 & 0.083 \\
 \rowcolor{light-gray}
 &  \textbf{64} & \textbf{3.142} & \textbf{0.02} & \textbf{7.525} & \textbf{0.087} & \textbf{20.195} & \textbf{0.094} & \textbf{0.938} & \textbf{0.001} & \textbf{0.811} & \textbf{0.005} & \textbf{121.804} & \textbf{0.164} & \textbf{0.085} \\ \hline
\multirow{6}{*}{\textbf{Passive Aggressive Regressor}} & 6 & 13.171 & 17.324 & 18.766 & 18.929 & 45.013 & 29.659 & 0.235 & 2.193 & 3.403 & 4.476 & 0.017 & 0.003 & 0.004 \\
 & 9 & 14.211 & 19.874 & 20.498 & 21.539 & 46.119 & 30.501 & 0.049 & 2.434 & 3.671 & 5.134 & 0.019 & 0.007 & 0.004 \\
 & 12 & 12.34 & 11.747 & 17.91 & 12.425 & 46.945 & 24.475 & 0.484 & 0.836 & 3.187 & 3.034 & 0.023 & 0.005 & 0.004 \\
 & 20 & 7.659 & 3.185 & 12.663 & 2.617 & 34.028 & 8.427 & 0.816 & 0.083 & 1.977 & 0.822 & 0.027 & 0.006 & 0.004 \\
 & 32 & 7.717 & 2.278 & 12.804 & 1.952 & 36.309 & 9.809 & 0.815 & 0.061 & 1.991 & 0.588 & 0.035 & 0.009 & 0.004 \\
 & \textbf{64} & \textbf{6.38} & \textbf{1.379} & \textbf{9.729} & \textbf{1.228} & \textbf{34.621} & \textbf{5.144} & \textbf{0.894} & \textbf{0.028} & \textbf{1.646} & \textbf{0.357} & \textbf{0.046} & \textbf{0.002} & \textbf{0.005} \\ \hline
\multirow{6}{*}{\textbf{SGD Regressor}} & 6 & 3.913 & 0.148 & 9.8 & 0.015 & 22.547 & 0.616 & 0.894 & 0.001 & 1.011 & 0.038 & 0.012 & 0.003 & 0.004 \\
 & 9 & 3.997 & 0.215 & 9.81 & 0.034 & 22.908 & 0.718 & 0.894 & 0.001 & 1.033 & 0.056 & 0.017 & 0.007 & 0.004 \\
 & 12 & 3.946 & 0.148 & 9.806 & 0.015 & 22.789 & 0.671 & 0.894 & 0.001 & 1.019 & 0.038 & 0.018 & 0.004 & 0.004 \\
 & \textbf{20} & \textbf{3.886} & \textbf{0.203} & \textbf{9.817} & \textbf{0.02} & \textbf{22.288} & \textbf{0.977} & \textbf{0.894} & \textbf{0.001} & \textbf{1.003} & \textbf{0.053} & \textbf{0.019} & \textbf{0.001} & \textbf{0.004} \\
 & 32 & 4.051 & 0.327 & 9.839 & 0.059 & 22.971 & 1.147 & 0.893 & 0.001 & 1.045 & 0.084 & 0.026 & 0.005 & 0.004 \\
 & 64 & 4.236 & 0.286 & 7.556 & 0.154 & 25.344 & 0.764 & 0.937 & 0.003 & 1.094 & 0.074 & 0.041 & 0.007 & 0.005 \\ \hline
\multirow{6}{*}{\textbf{LSTM}} & 6  & 5.001 & 0.104 & 11.007 & 0.07  & 25.666 & 0.284 & 0.866 & 0.002 & 1.292 & 0.027 & 14.475  & 0.006 & 0.066 \\
& 9  & 4.978 & 0.094 & 10.926 & 0.093 & 25.42  & 0.327 & 0.868 & 0.002 & 1.286 & 0.024 & 21.487  & 0.01  & 0.066 \\
& \textbf{12} & \textbf{4.811} & \textbf{0.098} & \textbf{10.765} & \textbf{0.081} & \textbf{24.574} & \textbf{0.411} & \textbf{0.872} & \textbf{0.002} & \textbf{1.243} & \textbf{0.025} & \textbf{28.351}  & \textbf{0.013} & \textbf{0.066} \\
& 20 & 4.865 & 0.101 & 10.694 & 0.082 & 24.566 & 0.449 & 0.874 & 0.002 & 1.256 & 0.026 & 45.884  & 0.045 & 0.066 \\
& 32 & 4.855 & 0.086 & 10.679 & 0.084 & 24.726 & 0.329 & 0.874 & 0.002 & 1.252 & 0.022 & 70.148  & 0.069 & 0.066 \\
& 64 & 4.853 & 0.084 & 10.675 & 0.086 & 24.777 & 0.334 & 0.875 & 0.002 & 1.253 & 0.022 & 166.131 & 0.18  & 0.066 \\ \hline
\multirow{6}{*}{\textbf{BI-LSTM}} & \textbf{6}  & \textbf{3.279} & \textbf{0.043} & \textbf{7.448} & \textbf{0.07}  & \textbf{19.899} & \textbf{0.592} & \textbf{0.939} & \textbf{0.001} & \textbf{0.847} & \textbf{0.011} & \textbf{279.032} & \textbf{0.178} & \textbf{0.131} \\
& 9  & 3.279 & 0.043 & 7.449 & 0.071 & 19.905 & 0.595 & 0.939 & 0.001 & 0.847 & 0.011 & 359.11  & 0.259 & 0.131 \\
& 12 & 3.28  & 0.043 & 7.45  & 0.071 & 19.911 & 0.599 & 0.939 & 0.001 & 0.847 & 0.011 & 408.813 & 0.257 & 0.131 \\
& 20 & 3.283 & 0.043 & 7.456 & 0.07  & 19.928 & 0.593 & 0.939 & 0.001 & 0.848 & 0.011 & 529.375 & 0.559 & 0.131 \\
& 32 & 3.286 & 0.043 & 7.461 & 0.07  & 19.952 & 0.592 & 0.939 & 0.001 & 0.848 & 0.011 & 716.442 & 0.774 & 0.131 \\
& 64 & 3.291 & 0.044 & 7.462 & 0.073 & 19.994 & 0.604 & 0.939 & 0.001 & 0.85  & 0.011 & 977.366 & 1.052 & 0.131 \\ \hline
\multirow{6}{*}{\textbf{Gated Recurrent Units}} & 6  & 5.047 & 0.117 & 10.71  & 0.12  & 25.807 & 0.264 & 0.873 & 0.003 & 1.304 & 0.03  & 32.276  & 0.028 & 0.049 \\
& 9  & 5.005 & 0.092 & 10.633 & 0.131 & 25.722 & 0.204 & 0.875 & 0.003 & 1.293 & 0.024 & 84.692  & 0.052 & 0.049 \\
& 12 & 5.004 & 0.107 & 10.55  & 0.107 & 25.775 & 0.233 & 0.877 & 0.003 & 1.292 & 0.028 & 102.017 & 0.045 & 0.049 \\
& \textbf{20} & \textbf{4.961} & \textbf{0.097} & \textbf{10.497} & \textbf{0.09}  & \textbf{25.648} & \textbf{0.219} & \textbf{0.878} & \textbf{0.002} & \textbf{1.281} & \textbf{0.025} & \textbf{76.746}  & \textbf{0.037} & \textbf{0.049} \\
& 32 & 4.978 & 0.091 & 10.507 & 0.091 & 25.793 & 0.241 & 0.878 & 0.002 & 1.284 & 0.023 & 108.012 & 0.09  & 0.049 \\
& 64 & 4.981 & 0.094 & 10.507 & 0.1   & 25.871 & 0.173 & 0.879 & 0.002 & 1.286 & 0.024 & 362.131 & 0.123 & 0.049 \\ \hline
\multirow{6}{*}{\textbf{LSTM with Attention}} & 6 & 13.373 & 3.154 & 19.175 & 3.599 & 42.255 & 6.426 & 0.581 & 0.135 & 3.455 & 0.815 & 1138.691 & 0.414 & 0.213 \\
& 9  & 9.537  & 4.483 & 15.36  & 4.769 & 34.566 & 9.085 & 0.716 & 0.169 & 2.464 & 1.158 & 1112.52  & 0.413 & 0.216 \\
& 12 & 8.46   & 4.342 & 13.988 & 4.717 & 32.983 & 9.179 & 0.761 & 0.172 & 2.185 & 1.122 & 1045.146 & 0.411 & 0.219 \\
& \textbf{20} & \textbf{7.258}  & \textbf{3.805} & \textbf{12.622} & \textbf{3.893} & \textbf{30.334} & \textbf{7.686} & \textbf{0.809} & \textbf{0.137} & \textbf{1.874} & \textbf{0.982} & \textbf{1083.556} & \textbf{0.423} & \textbf{0.227} \\
& 32 & 8.248  & 2.427 & 12.791 & 1.997 & 33.031 & 4.521 & 0.816 & 0.062 & 2.128 & 0.626 & 1122.657 & 0.41  & 0.239 \\
& 64 & 7.94   & 2.849 & 11.637 & 2.633 & 32.214 & 5.739 & 0.844 & 0.076 & 2.05  & 0.736 & 1052.156 & 0.397 & 0.27  \\ \hline
\multirow{6}{*}{\textbf{Linear Regression}} & 6 & 28.025 & 15.613 & 33.05 & 17.468 & 65.17 & 22.007 & -0.527 & 1.661 & 7.241 & 4.034 & 0.005 & 0.001 & 0.001 \\
 & 9 & 25.464 & 9.304 & 30.238 & 10.079 & 61.45 & 14.192 & -0.116 & 0.737 & 6.579 & 2.404 & 0.005 & 0.001 & 0.001 \\
 & 12 & 25.337 & 9.129 & 29.865 & 9.816 & 62.513 & 14.505 & -0.086 & 0.731 & 6.544 & 2.358 & 0.005 & 0.001 & 0.001 \\
 & 20 & 24.673 & 11.559 & 29.545 & 12.59 & 61.166 & 18.031 & -0.129 & 1.065 & 6.37 & 2.984 & 0.005 & 0.001 & 0.001 \\
 & 32 & 25.19 & 12.355 & 30.552 & 13.313 & 61.847 & 19.29 & -0.214 & 1.16 & 6.498 & 3.187 & 0.006 & 0.001 & 0.001 \\
 & \textbf{64} & \textbf{20.05} & \textbf{5.906} & \textbf{24.994} & \textbf{6.018} & \textbf{54.479} & \textbf{9.883} & \textbf{0.276} & \textbf{0.344} & \textbf{5.177} & \textbf{1.525} & \textbf{0.009} & \textbf{0.001} & \textbf{0.001} \\ \hline
\multirow{6}{*}{\textbf{SVR}} & 6 & 28.025 & 15.613 & 33.05 & 17.468 & 65.17 & 22.007 & -0.527 & 1.661 & 7.241 & 4.034 & 0.005 & 0.001 & 0.001 \\
 & 9 & 25.464 & 9.304 & 30.238 & 10.079 & 61.45 & 14.192 & -0.116 & 0.737 & 6.579 & 2.404 & 0.005 & 0.001 & 0.001 \\
 & 12 & 25.337 & 9.129 & 29.865 & 9.816 & 62.513 & 14.505 & -0.086 & 0.731 & 6.544 & 2.358 & 0.005 & 0.001 & 0.001 \\
 & 20 & 24.673 & 11.559 & 29.545 & 12.59 & 61.166 & 18.031 & -0.129 & 1.065 & 6.37 & 2.984 & 0.005 & 0.001 & 0.001 \\
 & 32 & 25.19 & 12.355 & 30.552 & 13.313 & 61.847 & 19.29 & -0.214 & 1.16 & 6.498 & 3.187 & 0.006 & 0.001 & 0.001 \\
 & \textbf{64} & \textbf{20.05} & \textbf{5.906} & \textbf{24.994} & \textbf{6.018} & \textbf{54.479} & \textbf{9.883} & \textbf{0.276} & \textbf{0.344} & \textbf{5.177} & \textbf{1.525} & \textbf{0.005} & \textbf{0.001} & \textbf{0.001} \\ \hline
\end{tabular}
}
\end{table}

Boxplots for WS 6, 9, and 12 behave similarly in terms of MAE. Consequently, only boxplots for WSs 6, 32, and 64 are then presented in Figure \ref{fig:box_plot_exp1}. The interquartile range (IQR) can define a consistent MAE, which helps assess model stability and variability. 

XGBoost and Random Forest consistently perform well across all window sizes (WSs), demonstrating low interquartile ranges (IQRs) and stable error rates. Similarly, BI-LSTM performs well but at a higher training and inference time and memory consumption. Larger window sizes tend to enhance stability in error rates across most models.

When the MAE is low and the RMSE is high in Table \ref{tab:exp1_full_table}, it indicates that while the average error is small, there are occasional large errors that cause the RMSE to be large. 
This is evident in models like support vector regression (SVR), linear regression (LR), and LSTM with attention. Indeed, LR and SVR are by far the worst-performing methods compared across all predictive error metrics. Conversely, models with very low RMSE, such as XGBoost, Adaboost, the default \textit{scikit-learn's} decision tree (CART), random forest (RF), stochastic gradient descent (SGD), LSTM, and BI-LSTM, indicate that their predictions do not tend to have spikes with large deviations from the ground truth. Hence being more reliable over time. 

Figure \ref{fig:exp1_scaterplots} depicts MAE obtained by models per WS over training times. 

Analyzing the top five best models in terms of MAE and training times, the SGD consistently achieves the shortest training time across all WSs. 
XGB offers comparable results to RF in terms of MAE and boasts a faster training and inference time across all WSs. LR and SVR are the fastest models overall during training time but with a high MAE. All of the algorithms ran in this experiment have low memory consumption and inference runtime footprints, thus being suitable for edge devices. 
The best overall models are XGBoost and RF, considering all factors: lowest errors, minimal training and inference times, and efficient memory usage. XGBoost offers the best balance between performance and resource consumption if training times are considered, making it ideal for resource-constrained applications that may need re-training at the edge, where low error rates and efficient use of time and memory are crucial. 
In terms of speed among the top models, SGD is the fastest for all window sizes. The Bidirectional LSTM has high training times in CPU but obtains low predictive errors, comparable to RF and XGBoost. Thus, it may be considered for edge devices with GPU built-in or when re-training does not need to occur in a timely manner and on the device.

\subsection{Experiment II}

The second experiment evaluates online learning algorithms, as outlined in Table \ref{tab:exp2_full_table}, using a \textit{prequential evaluation}. 
Online ML models are envisioned to learn on the fly, continuously adapting as new data arrives. Thus, during the evaluation of this experiment, we perform model updates~\cite{SUAREZCETRULO2023118934}. 
For more information about this process, refer to the source code in GitHub. 

\begin{table}[ht!]
\centering
\caption{Results for Experiment II, showcasing the best-performing model metrics WS with the lowest MAE across 20 runs. Values are rounded to a maximum of three decimal places.}
\label{tab:exp2_full_table}
\resizebox{1.0\columnwidth}{!}{%
\hskip-0.0cm\begin{tabular}{lllllllllllllll}
\hline
\multirow{2}{*}{\textbf{Model}} & \multirow{2}{*}{\textbf{Window Size}} & \multicolumn{2}{l}{\textbf{MAE}} & \multicolumn{2}{l}{\textbf{RMSE}} & \multicolumn{2}{l}{\textbf{SMAPE}} & \multicolumn{2}{l}{\(\mathbf{R^2}\)} & \multicolumn{2}{l}{\textbf{MASE}} & \textbf{Pretraining (s)} & \textbf{Evaluation (s)} & \textbf{Memory} \\ \cline{3-15} 
 &  & \textbf{mean} & \textbf{std} & \textbf{mean} & \textbf{std} & \textbf{mean} & \textbf{std} & \textbf{mean} & \textbf{std} & \textbf{mean} & \textbf{std} & \textbf{mean} & \textbf{mean} & \textbf{mean} \\ \hline
 \rowcolor{light-gray}
\multirow{6}{*}{\textbf{ARF}} & \textbf{6} & \textbf{3.427} & \textbf{0.018} & \textbf{9.078} & \textbf{0.023} & \textbf{20.17} & \textbf{0.131} & \textbf{0.909} & \textbf{0.0} & \textbf{0.885} & \textbf{0.005} & \textbf{71.431} & \textbf{31.641} & \textbf{146.031} \\
 & 9 & 3.553 & 0.038 & 9.212 & 0.066 & 20.291 & 0.162 & 0.906 & 0.001 & 0.918 & 0.01 & 95.346 & 40.418 & 184.937 \\
 & 12 & 3.729 & 0.08 & 9.429 & 0.141 & 20.721 & 0.204 & 0.902 & 0.003 & 0.963 & 0.021 & 99.159 & 41.007 & 187.451 \\
 & 20 & 4.193 & 0.177 & 9.952 & 0.256 & 21.949 & 0.445 & 0.891 & 0.006 & 1.083 & 0.046 & 113.908 & 42.076 & 157.025 \\
 & 32 & 5.63 & 0.765 & 11.28 & 0.756 & 25.978 & 1.812 & 0.859 & 0.019 & 1.452 & 0.197 & 140.108 & 44.412 & 93.727 \\
 & 64 & 11.661 & 0.213 & 17.159 & 0.24 & 38.946 & 0.481 & 0.676 & 0.009 & 3.011 & 0.055 & 153.104 & 38.984 & 3.986 \\ \hline
\multirow{6}{*}{\textbf{HAT Regressor}} & \textbf{6} & \textbf{3.795} & \textbf{0.063} & \textbf{9.34} & \textbf{0.085} & \textbf{21.583} & \textbf{0.42} & \textbf{0.904} & \textbf{0.002} & \textbf{0.981} & \textbf{0.016} & \textbf{4.567} & \textbf{2.575} & \textbf{2.819} \\
 & 9 & 3.924 & 0.062 & 9.454 & 0.108 & 22.128 & 0.311 & 0.901 & 0.002 & 1.014 & 0.016 & 5.786 & 2.886 & 4.411 \\
 & 12 & 4.039 & 0.085 & 9.61 & 0.203 & 22.626 & 0.624 & 0.898 & 0.004 & 1.043 & 0.022 & 7.075 & 3.185 & 5.815 \\
 & 20 & 4.27 & 0.134 & 9.679 & 0.183 & 23.467 & 0.609 & 0.897 & 0.004 & 1.102 & 0.035 & 10.8 & 4.007 & 10.013 \\
 & 32 & 6.198 & 3.352 & 11.59 & 3.85 & 28.057 & 6.95 & 0.836 & 0.134 & 1.599 & 0.865 & 17.498 & 5.569 & 10.367 \\
 & 64 & 9.533 & 3.835 & 14.139 & 4.543 & 35.126 & 7.241 & 0.759 & 0.162 & 2.462 & 0.99 & 41.728 & 11.697 & 8.146 \\ \hline
\multirow{6}{*}{\textbf{HT}} & \textbf{6} & \textbf{3.75} & \textbf{0.0} & \textbf{9.233} & \textbf{0.0} & \textbf{20.943} & \textbf{0.0} & \textbf{0.906} & \textbf{0.0} & \textbf{0.969} & \textbf{0.0} & \textbf{3.208} & \textbf{2.348} & \textbf{2.012} \\
 & 9 & 3.825 & 0.0 & 9.292 & 0.0 & 21.185 & 0.0 & 0.905 & 0.0 & 0.988 & 0.0 & 4.593 & 2.666 & 3.351 \\
 & 12 & 3.894 & 0.0 & 9.337 & 0.0 & 21.637 & 0.0 & 0.904 & 0.0 & 1.006 & 0.0 & 5.935 & 2.948 & 4.851 \\
 & 20 & 4.2 & 0.0 & 9.641 & 0.0 & 22.426 & 0.0 & 0.897 & 0.0 & 1.084 & 0.0 & 10.323 & 4.239 & 8.457 \\
 & 32 & 4.312 & 0.0 & 9.709 & 0.0 & 23.301 & 0.0 & 0.896 & 0.0 & 1.112 & 0.0 & 17.331 & 6.087 & 12.998 \\
 & 64 & 5.281 & 0.0 & 8.826 & 0.0 & 27.699 & 0.0 & 0.914 & 0.0 & 1.364 & 0.0 & 34.868 & 10.856 & 19.687 \\ \hline
\multirow{6}{*}{\textbf{SRP Regressor}} & \textbf{6} & \textbf{4.574} & \textbf{0.016} & \textbf{10.772} & \textbf{0.028} & \textbf{24.048} & \textbf{0.106} & \textbf{0.872} & \textbf{0.001} & \textbf{1.182} & \textbf{0.004} & \textbf{117.284} & \textbf{30.116} & \textbf{0.36} \\
 & 9 & 4.674 & 0.02 & 10.786 & 0.033 & 24.403 & 0.104 & 0.872 & 0.001 & 1.208 & 0.005 & 135.235 & 34.524 & 0.408 \\
 & 12 & 4.728 & 0.017 & 10.778 & 0.027 & 24.999 & 0.134 & 0.872 & 0.001 & 1.221 & 0.005 & 170.629 & 42.911 & 0.498 \\
 & 20 & 4.759 & 0.024 & 10.772 & 0.034 & 25.663 & 0.165 & 0.872 & 0.001 & 1.229 & 0.006 & 258.696 & 63.778 & 0.797 \\
 & 32 & 4.724 & 0.022 & 10.728 & 0.038 & 26.106 & 0.151 & 0.873 & 0.001 & 1.219 & 0.006 & 386.009 & 93.432 & 1.055 \\
 & 64 & 5.609 & 0.125 & 11.96 & 0.244 & 28.826 & 0.395 & 0.843 & 0.006 & 1.448 & 0.032 & 718.221 & 175.761 & 1.43 \\ \hline
\multirow{6}{*}{\textbf{PA}} & 6 & 8.987 & 0.054 & 18.224 & 0.053 & 38.455 & 0.154 & 0.633 & 0.002 & 2.322 & 0.014 & 0.002 & 3.747 & 0.003 \\
 & 9 & 8.927 & 0.035 & 17.781 & 0.036 & 37.917 & 0.127 & 0.651 & 0.001 & 2.306 & 0.009 & 0.003 & 3.759 & 0.004 \\
 & 12 & 8.892 & 0.061 & 17.617 & 0.12 & 37.679 & 0.265 & 0.657 & 0.005 & 2.297 & 0.016 & 0.003 & 3.748 & 0.004 \\
 & 20 & 9.501 & 0.056 & 17.651 & 0.089 & 38.875 & 0.188 & 0.656 & 0.003 & 2.453 & 0.015 & 0.004 & 3.755 & 0.004 \\
 & 32 & 10.13 & 0.057 & 17.684 & 0.07 & 40.04 & 0.172 & 0.655 & 0.003 & 2.613 & 0.015 & 0.005 & 3.768 & 0.004 \\
 & \textbf{64} & \textbf{6.764} & \textbf{0.017} & \textbf{10.395} & \textbf{0.014} & \textbf{34.72} & \textbf{0.059} & \textbf{0.881} & \textbf{0.0} & \textbf{1.747} & \textbf{0.004} & \textbf{0.01} & \textbf{3.739} & \textbf{0.004} \\ \hline
\multirow{6}{*}{\textbf{SGD Regressor}} & \textbf{6} & \textbf{3.897} & \textbf{0.008} & \textbf{9.814} & \textbf{0.001} & \textbf{21.661} & \textbf{0.007} & \textbf{0.894} & \textbf{0.0} & \textbf{1.007} & \textbf{0.002} & \textbf{0.002} & \textbf{3.73} & \textbf{0.004} \\
 & 9 & 3.906 & 0.009 & 9.824 & 0.001 & 21.616 & 0.016 & 0.893 & 0.0 & 1.009 & 0.002 & 0.003 & 3.739 & 0.004 \\
 & 12 & 3.911 & 0.006 & 9.829 & 0.001 & 21.742 & 0.082 & 0.893 & 0.0 & 1.01 & 0.002 & 0.003 & 3.737 & 0.004 \\
 & 20 & 3.951 & 0.006 & 9.846 & 0.001 & 22.15 & 0.041 & 0.893 & 0.0 & 1.02 & 0.002 & 0.004 & 3.735 & 0.004 \\
 & 32 & 4.015 & 0.012 & 9.861 & 0.002 & 22.494 & 0.036 & 0.893 & 0.0 & 1.036 & 0.003 & 0.005 & 3.743 & 0.004 \\
 & 64 & 4.201 & 0.01 & 7.7 & 0.003 & 25.323 & 0.036 & 0.935 & 0.0 & 1.085 & 0.003 & 0.01 & 3.717 & 0.004 \\ \hline
\multirow{6}{*}{\textbf{XGB regressor}} & 6 & 3.846 & 0.066 & 9.393 & 0.081 & 22.302 & 0.298 & 0.903 & 0.002 & 0.994 & 0.017 & 0.157 & 1173.884 & 0.004 \\
 & 9 & 3.898 & 0.076 & 9.41 & 0.071 & 22.279 & 0.319 & 0.902 & 0.001 & 1.007 & 0.02 & 0.143 & 1333.355 & 0.004 \\
 & 12 & 3.956 & 0.094 & 9.452 & 0.097 & 22.345 & 0.431 & 0.901 & 0.002 & 1.022 & 0.024 & 0.167 & 1481.934 & 0.004 \\
 & 20 & 4.032 & 0.103 & 9.462 & 0.094 & 22.788 & 0.532 & 0.901 & 0.002 & 1.041 & 0.027 & 0.223 & 1863.733 & 0.004 \\
 & 32 & 4.157 & 0.095 & 9.53 & 0.078 & 23.327 & 0.499 & 0.9 & 0.002 & 1.072 & 0.025 & 0.261 & 2411.73 & 0.004 \\ 
 \rowcolor{light-gray}
 &  \textbf{64} & \textbf{3.057} & \textbf{0.066} & \textbf{6.766} & \textbf{0.18} & \textbf{21.826} & \textbf{0.422} & \textbf{0.95} & \textbf{0.003} & \textbf{0.789} & \textbf{0.017} & \textbf{0.43} & \textbf{4271.722} & \textbf{0.004} \\ \hline
\end{tabular}%
}
\end{table}

\begin{figure}[ht!]
  \centering
  \includegraphics[width=\linewidth]{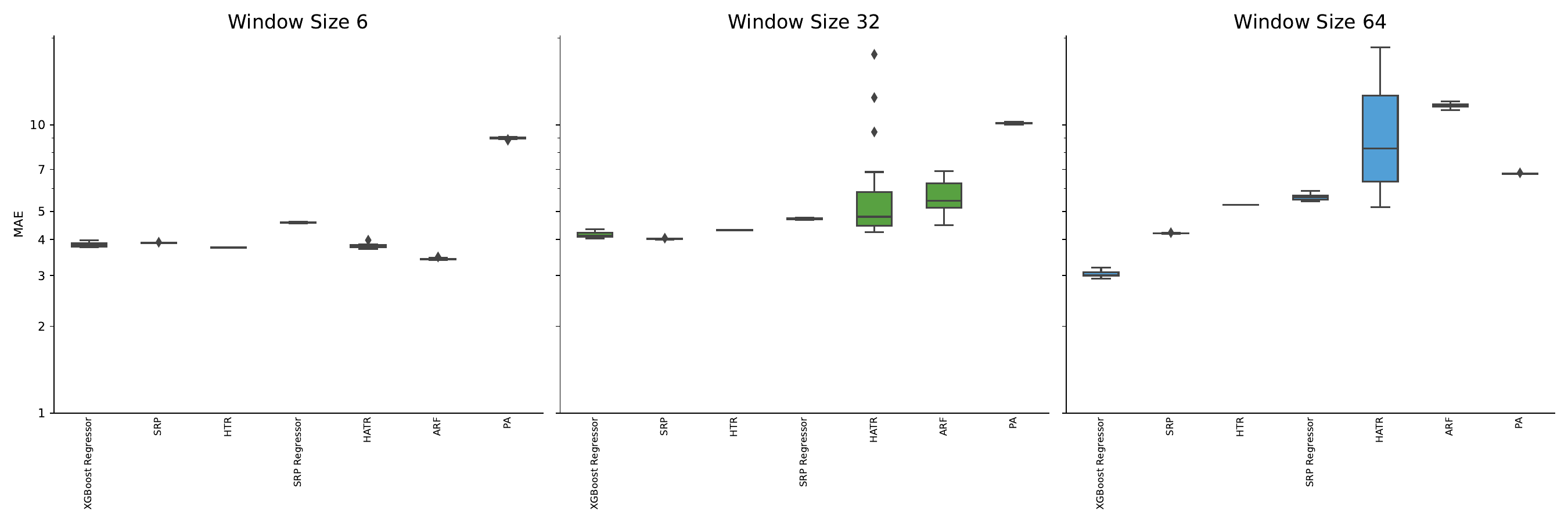}
    \caption{MAE per model in Experiment II at different window sizes.}
  \label{fig:mae_boxplot_exp2}
\end{figure}

\begin{figure}[ht!]
  \centering
  \includegraphics[width=\linewidth]{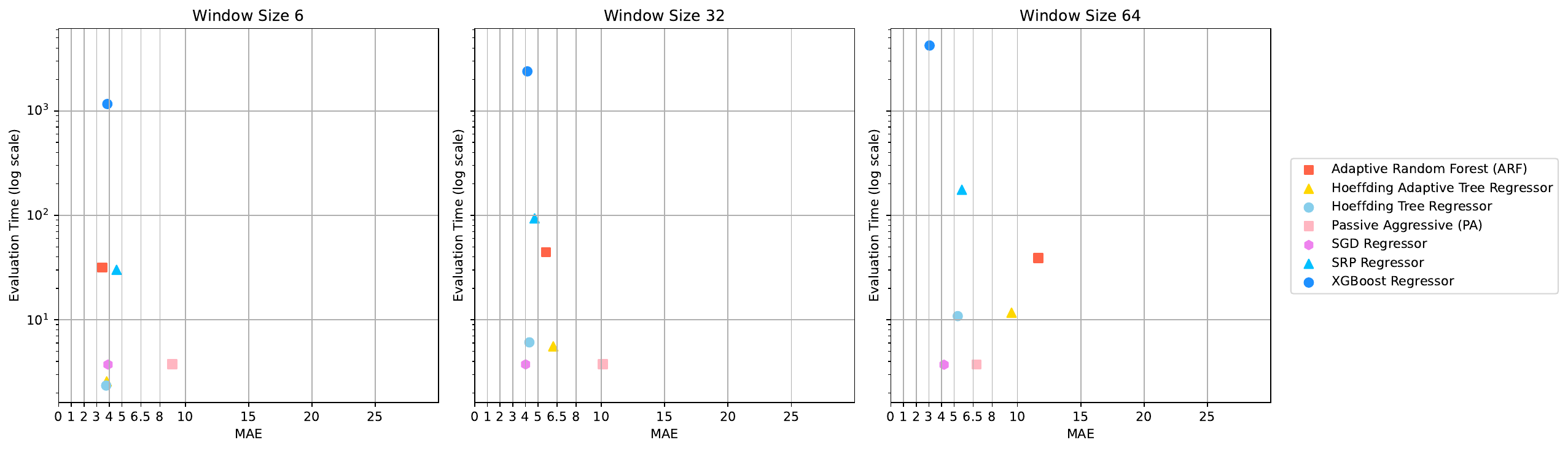}
    \caption{Prequential evaluation time vs. MAE per model in Experiment II.}
  \label{fig:exp2_scatter_p2}
\end{figure}

The algorithm with the best accuracies and lower training times in Experiment I was also evaluated prequentially in this experiment. Such evaluation entails continuous re-training, prequentially, after receiving each new data sample simulating a data stream for the XGBoost regressor, as this is not its incremental implementation. This is reflected in its evaluation time for Experiment II. 
The adaptive random forest algorithm and the two online DT algorithms (HAT and HT) also obtain low MAE values at WS 6 but exhibit higher memory usage and evaluation times. 
Despite its low pretraining time, XGBoost does not scale when being continuously retrained. 

Table \ref{tab:exp2_full_table} and Figure \ref{fig:mae_boxplot_exp2} show that XGBoosts overperforms all online learners in predictive accuracy. 

In (prequential) evaluation time, various models offer a good trade-off between MAE and efficiency (see Figure \ref{fig:exp2_scatter_p2}). 
Initially, HT excels in pre-training, evaluation, and memory usage for a window size 6. However, PA and SGD take the lead for larger window sizes 32 and 64. 
ARF obtained the second-best results in this experiment, although it underperformed offline ensembles in Experiment I.

\subsection{Experiment III}

In this experiment, we evaluate Lag-Llama in a similar setting to the previous experiments. The summary results for the window sizes (WS) with the lowest MAE are marked in bold in Table~\ref{tab:exp3_finetuned_lagllama}. 

Lag-Llama is evaluated there for zero-shot and four fine-tuned versions to understand the current state of time-series foundation models for evolving data streams. 
The original Lag-Llama implementation is primarily designed for forecasting single or multi-step-ahead predictions iteratively rather than for evaluating incoming data streams over time. 
To address this, we use each model's context length to represent the number of lags for each prediction. A data stream is then simulated over the evaluation set to perform a prequential evaluation for the fine-tuned version of Lag-Llama and compare results to Experiment II.

The experiment involved fine-tuning Lag-Llama models with lags of 32, 64, 128, and 256 (or window sizes) and also testing the same amounts as context lengths (\textit{CL}) in Lag-llama. Simultaneously, RoPE \cite{su2024roformer}, which utilizes rotatory positional embeddings (RoPE) scaling, is assessed. 
RoPE is evaluated to understand the relative position of lags within the series. 

Runtimes of Lag-Llama for fine-tuning range between [1070, 1850] seconds. 
The mean evaluation time ranged between [89, 107] seconds. 
All these times have been captured using a GPU (unlike Experiments I and II). Thus, it performs worse when compared to the previous experiments that were computed using the CPU. 

\begin{table}[ht!]
\centering
\caption{Results of Experiment III. Performance Comparison of MAE, SMAPE, MASE, and \( R^2 \)\ Metrics for Zero-shot and Fine-tuning Approaches on the CPU Dataset. Abbreviations – CL: context length}
\label{tab:exp3_finetuned_lagllama}
\resizebox{\columnwidth}{!}{%
\begin{tabular}{@{}lllllllllllll@{}}
\toprule
\multirow{2}{*}{\textbf{Model}} & \multirow{2}{*}{\textbf{CL}} & \multirow{2}{*}{\textbf{RoPE}} & \multicolumn{2}{l}{\textbf{MAE}} & \multicolumn{2}{l}{\textbf{RMSE}} & \multicolumn{2}{l}{\(\mathbf{R^2}\)} & \multicolumn{2}{l}{\textbf{SMAPE}} & \multicolumn{2}{l}{\textbf{MASE}} \\ \cmidrule(l){4-13} 
 &  &  & \textbf{mean} & \textbf{std} & \textbf{mean} & \textbf{std} & \textbf{mean} & \textbf{std} & \textbf{mean} & \textbf{std} & \textbf{mean} & \textbf{std} \\ \midrule
\multirow{8}{*}{\textbf{Zero shot}} & \multirow{2}{*}{32} & No & 6.252 & 0.016 & 14.587 & 0.088 & 0.753 & 0.002 & 26.692 & 0.043 & 2.248 & 0.014 \\
 &  & Yes & 6.249 & 0.010 & 14.566 & 0.015 & 0.753 & 0.001 & 26.661 & 0.054 & 2.247 & 0.015 \\
 & \multirow{2}{*}{64} & No & 9.819 & 0.020 & 19.089 & 0.027 & 0.583 & 0.001 & 35.981 & 0.117 & 3.237 & 0.015 \\
 &  & Yes & 9.355 & 0.017 & 18.848 & 0.021 & 0.576 & 0.001 & 32.656 & 0.071 & 3.231 & 0.017 \\
 & \multirow{2}{*}{128} & No & 8.847 & 0.021 & 15.142 & 0.041 & 0.737 & 0.001 & 37.356 & 0.138 & 2.129 & 0.013 \\
 &  & Yes & 7.112 & 0.019 & 14.304 & 0.037 & 0.771 & 0.001 & 33.109 & 0.065 & 1.759 & 0.010 \\
 & \multirow{2}{*}{256} & No & 9.651 & 0.021 & 14.304 & 0.044 & 0.747 & 0.001 & 39.567 & 0.148 & 1.949 & 0.011 \\
 &  & \textbf{Yes} & \textbf{5.500} & \textbf{0.021} & \textbf{11.579} & \textbf{0.034} & \textbf{0.857} & \textbf{0.001} & \textbf{32.021} & \textbf{0.169} & \textbf{1.169} & \textbf{0.004} \\ \midrule
\multirow{8}{*}{\textbf{Fine-tuned model on 32 lags}} & \multirow{2}{*}{32} & No & 5.393 & 0.694 & 10.651 & 0.488 & 0.844 & 0.020 & 24.775 & 1.091 & 2.106 & 0.306 \\
 &  & Yes & 5.271 & 0.645 & 10.703 & 0.742 & 0.844 & 0.025 & 24.460 & 0.783 & 2.037 & 0.238 \\
 & \multirow{2}{*}{64} & No & 4.941 & 0.623 & 8.482 & 0.792 & 0.905 & 0.020 & 24.432 & 1.005 & 1.558 & 0.323 \\
 &  & Yes & 4.967 & 0.708 & 8.439 & 0.867 & 0.906 & 0.022 & 24.370 & 1.211 & 1.463 & 0.386 \\
 & \multirow{2}{*}{128} & No & 4.184 & 0.441 & 7.204 & 0.502 & 0.937 & 0.010 & 23.652 & 0.959 & 1.128 & 0.108 \\
 &  & Yes & 4.114 & 0.322 & 7.274 & 0.344 & 0.937 & 0.006 & 23.504 & 0.571 & 1.095 & 0.091 \\
 & \multirow{2}{*}{256} & No & 3.623 & 0.324 & 7.074 & 0.277 & 0.942 & 0.005 & 22.585 & 0.758 & 0.904 & 0.078 \\
 &  & \textbf{Yes} & \textbf{3.567} & \textbf{0.150} & \textbf{7.053} & \textbf{0.211} & \textbf{0.942} & \textbf{0.004} & \textbf{22.615} & \textbf{0.586} & \textbf{0.905} & \textbf{0.040} \\ \midrule
\multirow{8}{*}{\textbf{Finetuned model on 64 lags}} & \multirow{2}{*}{32} & No & 5.184 & 0.517 & 10.548 & 0.466 & 0.851 & 0.016 & 24.852 & 0.937 & 1.849 & 0.142 \\
 &  & Yes & 5.383 & 0.631 & 10.850 & 0.733 & 0.841 & 0.022 & 24.566 & 1.062 & 2.063 & 0.304 \\
 & \multirow{2}{*}{64} & No & 4.838 & 0.644 & 8.345 & 0.548 & 0.910 & 0.013 & 24.469 & 1.101 & 1.362 & 0.235 \\
 &  & Yes & 4.778 & 0.675 & 8.132 & 0.726 & 0.912 & 0.020 & 24.722 & 1.093 & 1.599 & 0.289 \\
 & \multirow{2}{*}{128} & No & 3.817 & 0.283 & 6.945 & 0.342 & 0.942 & 0.007 & 22.968 & 0.528 & 1.094 & 0.111 \\
 &  & Yes & 3.881 & 0.329 & 7.233 & 0.647 & 0.937 & 0.013 & 23.021 & 0.705 & 1.073 & 0.104 \\
 \rowcolor{light-gray}
 & \multirow{2}{*}{256}    & \textbf{No} & \textbf{3.514} & \textbf{0.161} & \textbf{7.158} & \textbf{0.211} & \textbf{0.940} & \textbf{0.004} & \textbf{22.460} & \textbf{0.639} & \textbf{0.896} & \textbf{0.048} \\
 &  & Yes & 3.623 & 0.150 & 7.316 & 0.314 & 0.939 & 0.005 & 22.310 & 0.491 & 0.922 & 0.044 \\ \midrule
\multirow{8}{*}{\textbf{Finetuned model on 128 lags}} & \multirow{2}{*}{32} & No & 5.389 & 0.466 & 11.125 & 0.679 & 0.837 & 0.020 & 25.415 & 0.962 & 1.671 & 0.138 \\
 &  & Yes & 5.045 & 0.438 & 10.922 & 1.055 & 0.848 & 0.028 & 23.963 & 0.964 & 1.921 & 0.258 \\
 & \multirow{2}{*}{64} & No & 3.667 & 0.133 & 7.034 & 0.291 & 0.941 & 0.005 & 22.934 & 0.554 & 1.027 & 0.062 \\
 &  & Yes & 5.034 & 0.605 & 9.136 & 1.080 & 0.890 & 0.028 & 24.550 & 0.755 & 1.658 & 0.350 \\
 & \multirow{2}{*}{128} & No & 3.733 & 0.202 & 7.063 & 0.323 & 0.940 & 0.006 & 23.085 & 0.478 & 1.045 & 0.054 \\
 &  & Yes & 3.768 & 0.274 & 7.145 & 0.304 & 0.939 & 0.006 & 22.928 & 0.734 & 1.088 & 0.089 \\
 & \multirow{2}{*}{256} & No & 3.688 & 0.197 & 7.562 & 0.287 & 0.935 & 0.005 & 22.168 & 0.369 & 0.882 & 0.046 \\
 &  & \textbf{Yes} & \textbf{3.653} & \textbf{0.149} & \textbf{7.680} & \textbf{0.262} & \textbf{0.933} & \textbf{0.005} & \textbf{22.475} & \textbf{0.507} & \textbf{0.929} & \textbf{0.035} \\ \midrule
\multirow{8}{*}{\textbf{Finetuned model on 256 lags}} & \multirow{2}{*}{32} & No & 6.927 & 0.830 & 13.091 & 1.349 & 0.769 & 0.048 & 28.311 & 1.523 & 1.894 & 0.180 \\
 &  & Yes & 5.278 & 0.537 & 12.237 & 1.565 & 0.826 & 0.043 & 24.532 & 1.289 & 1.860 & 0.249 \\
 & \multirow{2}{*}{64} & No & 5.678 & 0.492 & 12.289 & 1.339 & 0.810 & 0.038 & 25.260 & 0.856 & 1.960 & 0.240 \\
 &  & Yes & 4.586 & 0.358 & 9.745 & 2.001 & 0.877 & 0.050 & 24.048 & 0.446 & 1.572 & 0.270 \\
 & \multirow{2}{*}{128} & No & 3.881 & 0.341 & 7.611 & 0.434 & 0.930 & 0.009 & 22.948 & 0.776 & 1.039 & 0.102 \\
 &  & Yes & 3.821 & 0.294 & 7.537 & 0.771 & 0.933 & 0.013 & 22.851 & 0.811 & 1.075 & 0.094 \\
 & \multirow{2}{*}{256} & No & 3.740 & 0.185 & 7.843 & 0.356 & 0.929 & 0.006 & 22.685 & 0.627 & 0.912 & 0.041 \\
 &  & \textbf{Yes} & \textbf{3.683} & \textbf{0.176} & \textbf{7.444} & \textbf{0.261} & \textbf{0.935} & \textbf{0.004} & \textbf{22.872} & \textbf{0.462} & \textbf{0.971} & \textbf{0.030} \\ \bottomrule
\end{tabular}%
}
\end{table}

From the results observed in Experiment III (see Table ~\ref{tab:exp3_finetuned_lagllama}), it is clear that, in comparison to Experiment II, none of the Lag-Llama tests in our study (neither the zero-shot nor the fine-tuned) were able to outperform ARF or the re-trained XGBoost from Experiment II.

\subsection{Discussion} 

In this work, choosing the best model across experiments involves a trade-off between performance and computational time. Finding the best model depends on the need for model updates and constraints in the devices needing to predict such workloads. In this study, models are targeted for constrained devices that need low computational inference times.  

In Experiment I, despite RF showing the best performance, XGBoost obtains very similar results at a lower computational cost. 

In Experiment II, XGBoost exhibits the highest predictive performance, although it comes with a considerable evaluation time, allowing ARF to take the lead in terms of performance metrics. Nevertheless, ARF still shows a relatively high memory consumption, although this should not be a concern for many edge device setups. 

In this work, ensemble models have shown the best overall predictive accuracies and the best tradeoff to computational cost. Online learners in Experiment II have still been able to compete with results from Experiment I but have not been able to overperform them. 
Models from Experiment II require fewer computational resources compared to deep learning methods in Experiment I or Lag-Llama in Experiment III, which will perform well at the edge when having access to GPU resources. 
While Lag-Llama is trained on extensive context lengths, it may encounter difficulties in accurately adapting to changes in evolving streams. Furthermore, the algorithms tested in Experiment I consistently outperform Lag-Llama, which largely mirrors the performance of online learners in Experiment II (ARF, HAT, HT) but with larger runtimes. 
Despite its low pretraining time in Experiment II and the fact that it obtained the best predictive accuracy in our experiments, XGBoost does not scale when being continuously retrained. This is understandable as the algorithm has not been designed for this purpose, and running an adaptive version should be a future line of work. As far as we know, this has not yet been implemented in the software (\textit{River}); hence, this work is out of our scope. 

In summary, online learners offer promising results, and a more in-depth study adding extra algorithms and hyperparameters may help find an optimal method. In the meantime, ensembles in Experiment I seem to be the best option for predicting CPU loads in edge devices. 
A more extensive study using data stream learning benchmarks should be made for this purpose, but it is considered out of scope in this work.

\section{Conclusion}\label{sec:conclusion}

This paper has presented an approach to predicting CPU utilization and allowing model selection between state-of-the-art, online ML methods and the time-series foundation model Lag-Llama. 
The results show promising results for online ML methods and underscore the use of non-linear methods like ensembles or neural networks in case of having access to GPUs to predict CPU load at the edge. 
The results obtained enforce the relevance of the dataset generated for data stream learning. 

Our study highlights the effectiveness of online ML methods as a suitable approach for CPU performance estimation. 
Further research is encouraged to explore additional applications and extend the proposed evaluation framework to other domains.

\section*{Acknowledgments}

Our advancement has been made possible by funding from the European Union's HORIZON research and innovation program (Grant No. 101070177).

We are grateful to our colleagues at the EU Horizon project ICOS and CeADAR (Ireland's National Centre for Applied AI) for helping to start and shape this research effort.

 \bibliographystyle{elsarticle} 
 \bibliography{references}

\end{document}